\pdfoutput=1

\documentclass[11pt]{article}

\usepackage[final]{ACL2023}

\usepackage{times}
\usepackage{xcolor}
\usepackage{booktabs}
\usepackage{multirow}
\usepackage{latexsym}
 \usepackage{graphicx}
 \usepackage{amsmath}
  \usepackage{booktabs}
\usepackage[T1]{fontenc}

\usepackage[utf8]{inputenc}

\usepackage{microtype}

\usepackage{inconsolata}

\title{Temporalizing Confidence:\\ Evaluation of Chain-of-Thought Reasoning with Signal Temporal Logic }

\author{Zhenjiang Mao\textsuperscript{1}, Artem Bisliouk\textsuperscript{1 2}, Rohith Reddy Nama\textsuperscript{1}, Ivan Ruchkin\textsuperscript{1}  \\
  University of Florida\textsuperscript{1}, University of Mannheim\textsuperscript{2} \\
  \texttt{\{z.mao,a.bisliouk,namarohithreddy,iruchkin\}@ufl.edu}}

\begin{document}
\maketitle
\begin{abstract}
\looseness=-1
Large Language Models (LLMs) have shown impressive performance in mathematical reasoning tasks when guided by Chain-of-Thought (CoT) prompting. However, they tend to produce highly confident yet incorrect outputs, which poses significant risks in domains like education, where users may lack the expertise to assess reasoning steps. To address this, we propose a structured framework that models stepwise confidence as a temporal signal and evaluates it using Signal Temporal Logic (STL). In particular, we define formal STL-based constraints to capture desirable temporal properties and compute robustness scores that serve as structured, interpretable confidence estimates. Our approach also introduces a set of uncertainty reshaping strategies to enforce smoothness, monotonicity, and causal consistency across the reasoning trajectory. Experiments show that our approach consistently improves calibration metrics and provides more reliable uncertainty estimates than conventional confidence aggregation and post-hoc calibration.
\end{abstract}

\section{Introduction}

Large language models (LLMs) are increasingly applied in educational contexts such as concept explanation, question answering, and personalized tutoring, especially in STEM domains like mathematics~\cite{kasneci2023chatgpt}. These models exhibit strong capabilities in solving complex problems; however, they also tend to produce answers that are fluent and seemingly confident, yet factually incorrect. In educational settings, such outputs can be particularly problematic, as students may lack the expertise to distinguish between correct and incorrect reasoning~\cite{polyxeni2024building}, and may be misled by responses that appear trustworthy. This mismatch between confidence and correctness raises critical concerns about the reliability of LLM-generated answers and highlights the importance of integrating uncertainty estimation into educational AI systems.

Although prior work has explored various uncertainty estimation techniques, such as predictive entropy, sampling-based variance, and confidence calibration, most studies focus on general NLP tasks rather than educational scenarios~\cite{zhao2021calibrate,jiang2021can}. In educational contexts like mathematics learning, well-calibrated uncertainty can be especially useful for guiding student attention, supporting teacher oversight, and improving feedback systems. However, existing uncertainty metrics often show poor alignment with actual correctness~\cite{zhu2025uncertainty}.

In this work, we address this challenge by proposing a novel approach to \textit{estimate uncertainty in LLM-based chain-of-thought (CoT) reasoning} for high school mathematics problems. 
Our method models the sequence of reasoning steps as a temporal confidence signal and evaluates its structural properties using Signal Temporal Logic (STL)~\cite{fainekos2006robustness}. Instead of modifying the LLM or directly penalizing its outputs, we quantify undesirable confidence behaviors, such as abrupt increases following uncertain steps, by computing robustness scores against formal STL constraints. This yields a constraint-aware aggregation scheme that captures how confidence is expected to evolve over time, offering a structured and interpretable view of the reasoning process while improving calibration.
We evaluate our method on a curated dataset of Chinese Gaokao mathematics multiple-choice questions~\cite{zhang2023evaluating}. Experimental results show that our method significantly improves calibration, reducing Expected Calibration Error (ECE) compared to baseline uncertainty aggregation methods. 

\begin{figure*}
	\centering         \includegraphics[width=2\columnwidth]{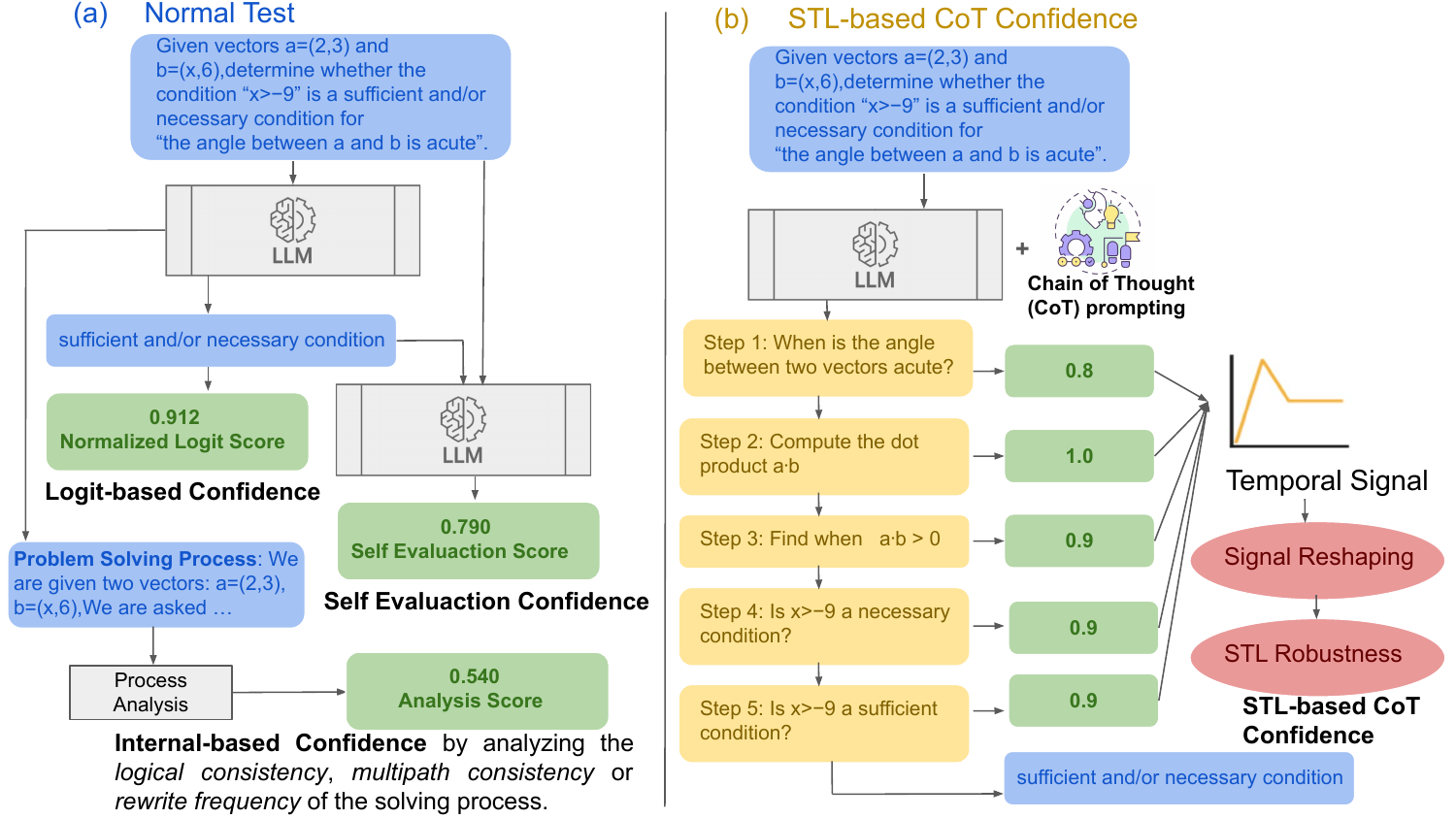}
\caption{(a) Conventional methods output global confidence via logit, self-evaluation, or internal analysis. (b) Our method models step-wise CoT confidence as a temporal signal, applies signal reshaping, and evaluates robustness using STL to obtain a temporally consistent confidence score.}
\label{fig:method_diagram}

 \end{figure*}

This paper's contributions are: (1) a novel perspective that treats stepwise confidence in chain-of-thought reasoning as a temporal signal amenable to formal analysis, (2) a constraint-aware modeling approach that reshapes confidence trajectories and quantifies their structural quality using STL robustness and (3) empirical validation of our method’s effectiveness on Chinese Gaokao mathematics. Fig~\ref{fig:method_diagram} provides an overview of our pipeline: starting from stepwise CoT confidence, we apply uncertainty reshaping followed by STL-based temporal logic evaluation. This transformation results in interpretable, structure-aware confidence scores.

\section{Related Work}
\textbf{Applications of LLMs in Education:} 
LLMs have been widely adopted in educational settings for tasks such as grammar correction, content generation, problem explanation, and intelligent tutoring. Prior studies highlight their potential to support learners across domains like language writing~\cite{kasneci2023chatgpt}, mathematics~\cite{gan2023large}, and personalized feedback~\cite{zhou2025study,wang2024large}. For instance, LLMs like MathGPT and Khanmigo have been used to generate step-by-step math explanations aligned with curriculum standards~\cite{shah2024ai}, while ChatGPT has shown promise in automated feedback for student essays and short answers~\cite{kasneci2023chatgpt}. Despite these advances, concerns remain around academic integrity, hallucinated outputs, and students over-relying on unverified responses ~\cite{benitez2024harnessing}. Moreover, few works explicitly quantify how reliable these educational outputs are, or how uncertainty signals can be used to guide learners or inform teachers. As LLMs become integral to education technology, recent surveys have called for deeper investigations into trust, transparency, and uncertainty in educational applications~\cite{idris2024revolutionising}.

\smallskip
\noindent \textbf{Uncertainty in LLMs:} 
LLMs exhibit remarkable fluency across diverse NLP tasks, yet their outputs often suffer from overconfidence and miscalibration~\cite{kadavath2022language}, especially in impact-sensitive domains such as education. Existing work has explored various uncertainty estimation techniques, including entropy-based methods like predictive entropy and confidence gaps~\cite{zhu2025uncertainty}, as well as sampling consistency across outputs~\cite{lyu2025calibrating}. Confidence calibration is another active area, revealing that LLMs tend to be overconfident, particularly in zero-shot or out-of-domain tasks~\cite{desai2020calibration,zhao2021calibrate}. Recent studies also propose using uncertainty signals to guide reasoning, such as Uncertainty-Guided CoT prompting~\cite{zhu2025uncertainty}, active prompting for data selection, and consistency-based calibration~\cite{diao2023active,lyu2025calibrating}. However, most of these methods have been evaluated on general NLP or code generation tasks, with limited attention to structured educational settings like math problem solving, where reliable uncertainty estimates can help students assess model-generated reasoning and assist teachers in diagnosing student understanding..

\smallskip
\noindent \textbf{CoT and STL}: 
Recent advances in CoT prompting have significantly improved the multi-step reasoning ability of LLMs, yet they also introduce new layers of uncertainty, such as error propagation across intermediate steps and unfaithful explanations~\cite{zhang2022automatic,wang2022self,tanneru2024quantifying}. In this work, we propose a novel perspective that treats CoT steps as discrete temporal signals, enabling the use of STL to formally specify and evaluate reasoning quality over time~\cite{rescher2012temporal}. STL allows for expressive specifications like \emph{eventually correct} or \emph{always consistent}, and provides quantitative robustness scores that capture the degree of satisfaction or violation~\cite{fainekos2006robustness}. This formalism has been successfully applied in domains such as motion planning~\cite{van2024specification}, reinforcement learning~\cite{li2017reinforcement}, and control synthesis, and offers a promising path toward interpretable and rigorous evaluation of LLM-generated reasoning trajectories. Applying STL to CoT would not only enable structured detection of flawed reasoning patterns but also facilitate the development of confidence-aware feedback and scoring systems in education and other applications.

\section{STL-Guided Confidence Estimation}

Our approach consists of three stages: (1) generating a stepwise confidence signal via CoT prompting, (2) applying uncertainty reshaping strategies to promote temporal consistency, and (3) evaluating the reshaped sequence using STL. This pipeline is illustrated in Fig~\ref{fig:method_diagram}(b), showing how each reshaping strategy transforms a sample confidence trajectory. Compared to the original signal, our smoothing strategies effectively suppress abrupt spikes and produce a more temporally coherent confidence trajectory, while preserving the overall trend of reasoning. This behavior is crucial for downstream STL-based evaluation, which benefits from smoother and more causally consistent input signals.

\begin{table*}[t]
\centering
\small
\begin{tabular}{|p{11cm}|c|c|c|}
\hline
\textbf{Question} & \textbf{Logit} & \textbf{Self-Eval} & \textbf{Internal} \\
\hline
Given vectors \( \mathbf{a} = (2, 3) \) and \( \mathbf{b} = (x, 6) \). Determine whether the condition “\( x > -9 \)” is a sufficient and/or necessary condition for “the angle between \( \mathbf{a} \) and \( \mathbf{b} \) is acute.” \newline
Options: A. Sufficient but not necessary \quad B. Necessary but not sufficient \quad C. Sufficient and necessary \quad D. Neither sufficient nor necessary \newline
\textit{Model answer: C (incorrect), True answer: \textbf{B}} & 
0.99 & 
0.98 & 
0.99 \\
\hline
Set parabola \( C: y^2 = 4x \). The focus is \( F \). A line passes through \( (-2, 0) \) with slope \( \frac{2}{3} \), intersecting \( C \) at points \( M \) and \( N \). Compute \( \vec{FM} \cdot \vec{FN} \). \newline
Options: A. 5 \quad B. 6 \quad C. 7 \quad \textbf{D. 8 (correct)} \newline
\textit{Model answer: D (correct)} & 
0.98 & 
0.95 & 
0.97 \\
\hline
\end{tabular}
\caption{Examples of high school mathematics questions and confidence scores from three estimation strategies: Logit-based, Self-evaluation, and Internal consistency. While the model is highly confident in all three views, the first question is incorrectly answered, leading to significant miscalibration. The \textbf{Expected Calibration Error (ECE)} for Logit, Self-Eval, and Internal confidences are \textbf{0.485}, \textbf{0.465}, and \textbf{0.480}, respectively.}
\label{tab:math_examples}
\end{table*}


\subsection{Problem Setup}

We model LLMs as autonomous agents tasked with solving high school mathematics problems. Given an input question \( q \in \mathcal{Q} \), the agent generates a final answer \( a \in \mathcal{A} \) along with a scalar uncertainty score \( u \in [0, 1] \), representing its confidence in the answer. Ideally, high confidence should correspond to high correctness probability, and vice versa~\cite{guo2017calibration}.

To evaluate calibration, we employ the \textit{Expected Calibration Error (ECE)}~\cite{naeini2015obtaining,desai2020calibration}, a widely-used metric that quantifies the mismatch between confidence and accuracy. Formally, we partition predictions into \( M \) bins based on their confidence values and compute:
\begin{equation}
    \mathrm{ECE} = \sum_{m=1}^{M} \frac{|B_m|}{n} \left| \mathrm{acc}(B_m) - \mathrm{conf}(B_m) \right|,
\end{equation}
where \( B_m \) denotes the set of predictions falling into the \( m \)-th confidence bin,   \( \mathrm{acc}(B_m) \) is the empirical accuracy defined as 
\begin{equation}
\mathrm{acc}(B_m) = \frac{1}{|B_m|} \sum_{i \in B_m} \mathbf{1}[\hat{a}_i = a_i],
\end{equation}
and \( \mathrm{conf}(B_m) \) is the average predicted confidence. Here, \( n \) is the total number of examples, and \( \mathbf{1}[\cdot] \) is the indicator function that returns 1 if the prediction is correct, and 0 otherwise. 
Since the task is formulated as multiple-choice classification, both model predictions and ground-truth answers are represented as one of a finite set of discrete options (e.g., A, B, C, D). This allows correctness to be determined via exact match of the selected option label, avoiding ambiguities arising from natural language variation.
rect, and 0 otherwise.


In addition to ECE, we also report the \textit{Brier Score (BS)} as a complementary calibration metric. The Brier Score measures the mean squared error between predicted confidence and ground-truth correctness, defined as:
\begin{equation}
    \mathrm{BS} = \frac{1}{n} \sum_{i=1}^{n} (c_i - y_i)^2,
\end{equation}
where \(c_i \in [0,1] \) is the model's predicted confidence for example \( i \), and \( y_i \in \{0,1\} \) is the binary correctness label. Lower Brier Scores indicate better calibrated and more reliable confidence estimates.

Unlike conventional classification tasks, reasoning in LLMs often unfolds over multiple steps~\cite{wei2022chain}. This raises an additional challenge: confidence should not only be calibrated across examples, but also evolve smoothly and consistently over the reasoning trajectory~\cite{zhu2025uncertainty}. Hence, the problem extends to generating temporally coherent uncertainty sequences that reflect both local confidence (per step) and global correctness (final answer). Our objective is thus to design a framework where uncertainty estimates are not only well-calibrated across examples, but also evolve in a temporally consistent manner—exhibiting properties such as smooth progression, causal coherence, and alignment with the underlying reasoning process.

\subsection{Uncertainty Reshaping Strategies}

To model reasoning-time uncertainty, we use CoT prompting to elicit a sequence of intermediate reasoning steps \( \{s_1, \dots, s_T\} \), each associated with a confidence score \( c_t \in [0,1] \). We treat the resulting confidence sequence \( \mathbf{c} = \{c_1, \dots, c_T\} \) as a temporal signal~\cite{rescher2012temporal}. However, due to the inherent causal nature of reasoning, abrupt increases in confidence, especially after initially low-confidence steps, can be misleading~\cite{zhu2025uncertainty}. To take advantage of this insight, we propose several signal reshaping functions that induce smoother and causally consistent confidence evolution. While some of these strategies are conceptually related to smoothing methods in time-series analysis, they are, to the best of our knowledge, novel in the context of modeling stepwise confidence in LLM-based reasoning.

\begin{figure*}
	\centering         \includegraphics[width=2\columnwidth]{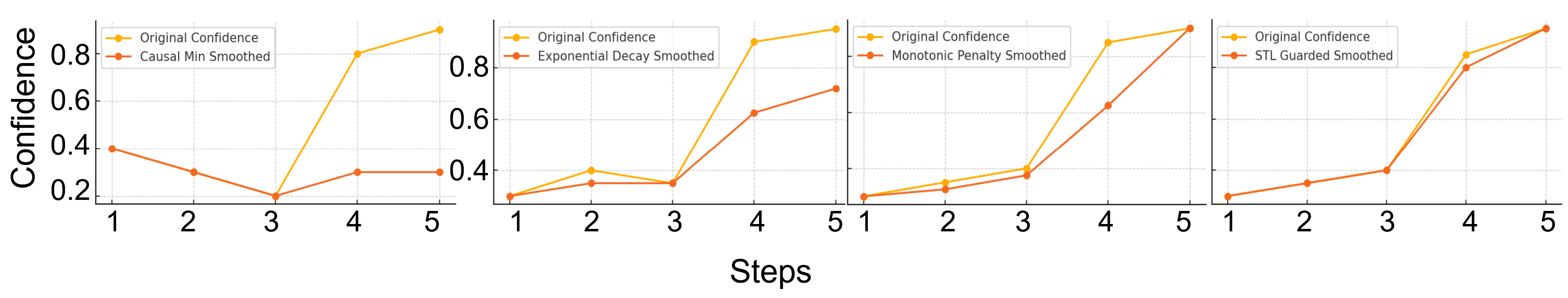}
\caption{Visualization examples of stepwise confidence signals before and after applying different uncertainty reshaping strategies. Each subplot compares the original confidence trajectory (yellow) against a smoothed version (orange) using one of the following methods: (left to right) Causal Minimum Smoothing (CMS), Exponential Decay Smoothing (EDS), Monotonic Penalty Smoothing (MPS), and Guarded Smoothing (GS). These transformations produce smoother and more temporally consistent confidence profiles while preserving the overall trend.}
\label{fig:smooth}
 \end{figure*}
\begin{itemize}
    \item \textbf{Causal Minimum Smoothing (CMS)}: Limits future confidence based on past minimum values plus a small fixed margin \( \delta \):
    \[
    \tilde{c}_t = \min\left(c_t, \min_{i < t} c_i + \delta\right)
    \]
    
    \item \textbf{Exponential Decay Smoothing (EDS)}: Applies exponential smoothing  by blending the current value with the average of past values:
    \[
    \tilde{c}_t = \alpha \cdot c_t + (1 - \alpha) \cdot \frac{1}{t} \sum_{i=1}^{t-1} c_i
    \]

    \item \textbf{Monotonic Penalty Smoothing (MPS)}: Dampens confidence spikes if the previous step is below a fixed threshold \( \tau \). This focuses on upward spikes, which are more likely to mislead following uncertain steps:
    \[
    \tilde{c}_t =
    \begin{cases}
        \dfrac{c_{t-1} + c_t}{2} & \text{if } c_{t-1} < \tau \text{ and } c_t > c_{t-1} \\
        c_t & \text{otherwise}
    \end{cases}
    \]

    \item \textbf{ Guarded Smoothing (GS)}: Caps sudden jumps beyond threshold \( \tau \) plus tolerance \( \epsilon \):
    \[
    \tilde{c}_t =
    \begin{cases}
        \tau + \epsilon, & \text{if } c_{t-1} < \tau \text{ and } c_t > \tau + \epsilon \\
        c_t, & \text{otherwise}
    \end{cases}
    \]
\end{itemize}

The reshaped sequence \( \tilde{\mathbf{c}} = \{\tilde{c}_1, \dots, \tilde{c}_T\} \) is passed to a formal temporal logic evaluation module described next as shown in Fig.~\ref{fig:method_diagram}(b).

\subsection{STL-Based Temporal Evaluation}

Rather than relying solely on the final-step confidence \( c_T \) or averaging all stepwise confidences, we propose a STL-based framework to evaluate the temporal structure of the confidence trajectory \( \tilde{\mathbf{c}} = \{\tilde{c}_1, \dots, \tilde{c}_T\} \)~\cite{fainekos2006robustness}. STL enables formal specification of desired temporal properties of confidence during reasoning, such as smooth progression or eventual certainty. 

Each STL formula encodes a specific temporal pattern, and its associated robustness score \( \rho_i = \rho(\tilde{\mathbf{c}}, \text{STL}) \in \textbf{R} \) quantifies how well the reshaped signal satisfies that property~\cite{donze2010robust}. A positive score indicates the satisfaction margin, while a negative score represents the magnitude of a violation.

We define three STL specifications, each yielding a separate confidence score:
\begin{itemize}
    \item \textbf{Eventually Confident:} Confidence should eventually rise above a threshold \( \tau \):
    \[
    \begin{aligned}
         \text{STL1} &= \Diamond_{[t_1, t_2]} (\tilde{c}(t) > \tau) \\
    \end{aligned}
    \]

    \item \textbf{Always Stable or Increasing:} Confidence should not drop abruptly:
    \[
    \begin{aligned}
\Delta\tilde{c}(t) &= \tilde{c}(t) - \tilde{c}(t-1) \\
 \text{STL2} &= \Box_{[t_{1},\,t_{2}]}\bigl(\,\Delta\tilde{c}(t) \ge -\epsilon\,\bigr) \\
    \end{aligned}
    \]

    \item \textbf{Local Smoothness:} Confidence should not change too much between steps:
    \[
    \begin{aligned}
             \text{STL3} &= \Box_{[t_{1},\,t_{2}]}\bigl(\,\lvert\Delta\tilde{c}(t)\rvert \le \delta\,\bigr)
    \end{aligned}
    \]

\end{itemize}

Each resulting score $\hat{c}  = \text{ReLU}(\rho(\tilde{\mathbf{c}}, \text{STL})) \in [0, 1] $ represents an interpretable, temporally-informed confidence score derived from logic-based robustness. These scores can be used independently for analysis or combined in multi-dimensional calibration evaluation.

While our STL-based scoring framework does not require labeled data during reasoning, it does rely on threshold hyperparameters (e.g., \( \tau, \epsilon, \delta \)) that influence robustness computation. To set them, we perform a grid search on a held-out validation set. This makes our approach partially \textit{post-hoc} in nature: only the STL evaluation stage requires data-driven tuning, whereas the preceding confidence reshaping is fully unsupervised and model-agnostic.

\begin{figure*}
	\centering         \includegraphics[width=2\columnwidth]{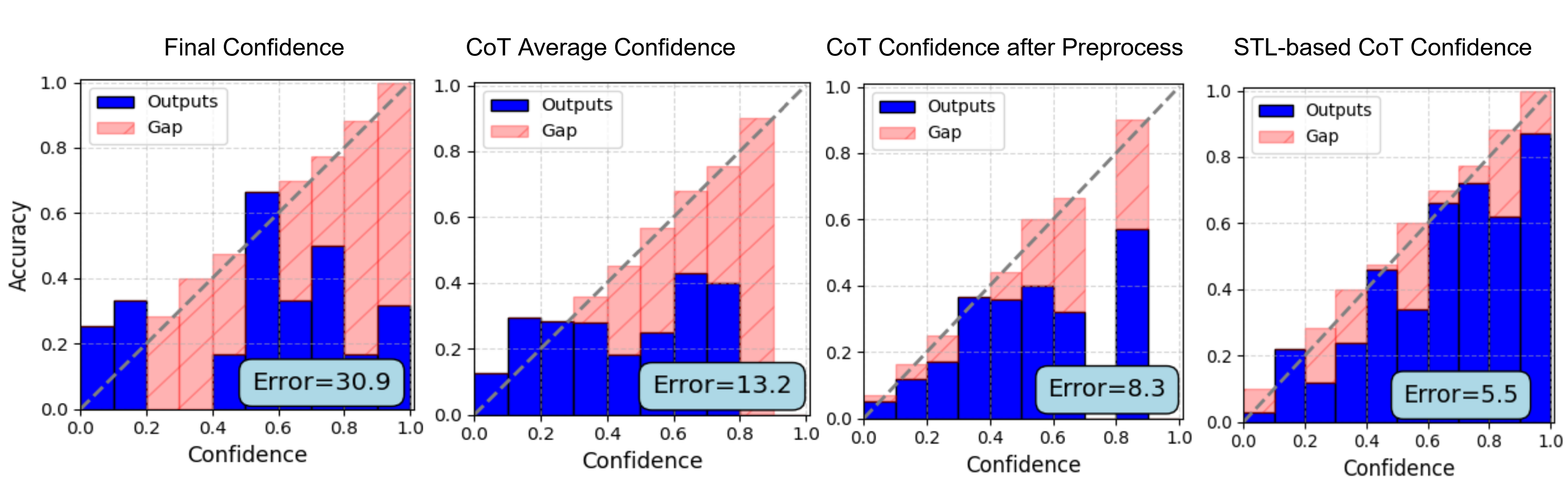}
\caption{ECE comparison of four confidence estimation methods. (1) Final-step confidence; (2) average confidence over CoT steps; (3) CoT confidence after applying Uncertainty Reshaping Strategies; and (4) STL-based CoT confidence, which combines Uncertainty Reshaping Strategies with the \textbf{STL1} formula (\textit{Eventually Confident:} confidence should eventually exceed a threshold \( \tau \)). The STL1-enhanced method achieves the best calibration (ECE = 5.5).}

\label{fig:results}

 \end{figure*}
 
\section{Experiments}
In Section 4, we present an ablation study and a comparison against established post-hoc calibration techniques that investigates the impact of STL parameterization and compares our method against established post-hoc calibration techniques such as \textit{Temperature Scaling}~\cite{guo2017calibration} and \textit{Histogram Binning}~\cite{zadrozny2001obtaining}. Our results show that STL-based evaluation provides not only competitive calibration performance but also interpretable, temporally grounded diagnostics of reasoning quality.

\begin{table*}[h]
\centering
\small
\begin{tabular}{|c|l|c|c|c|}
\hline
\textbf{Method} & \textbf{ Reshaping Strategy} & \textbf{Logits-based $\downarrow$} & \textbf{Self-evaluation-based  $\downarrow$} & \textbf{Internal-based $\downarrow$} \\
\hline
 1-step Uncertainty&- & $0.324 \pm 0.045$ & $0.692 \pm 0.035$ & $0.694 \pm 0.033$

\\
\hline

Temperature Scaling&- & $0.246 \pm 0.061$ & $0.158 \pm 0.033$ & $0.173 \pm 0.046$

\\
\hline
Histogram Binning&- & $0.139 \pm 0.004$ & $0.095 \pm 0.069$ & $0.185 \pm 0.129$

\\
\hline
\multirow{5}{*}{CoT Average} &- & $0.141 \pm 0.062$ & $0.542 \pm 0.039$ & $0.603 \pm 0.037$
\\
&CMS& \textbf{0.107 ± 0.048} & $0.486 \pm 0.040$ & $0.579 \pm 0.041$

\\

&EDS&$0.126 \pm 0.022$ & $0.502 \pm 0.036$ & $0.573 \pm 0.039$

\\

&MPS&$0.129 \pm 0.063$ & $0.530 \pm 0.037$ & $0.602 \pm 0.038$

\\

&GS&$0.140 \pm 0.060$ & $0.538 \pm 0.038$ & $0.579 \pm 0.021$

\\
\hline
\multirow{5}{*}{STL1 (Eventually Confident)} 
& -                     &$0.174 \pm 0.019$ & $0.082 \pm 0.021$ & $0.102 \pm 0.055$
 \\
& CMS            &     $0.250 \pm 0.198$ & $0.136 \pm 0.006$ & $0.119 \pm 0.023$

\\
& EDS                     & $0.236 \pm 0.064$ & $0.098 \pm 0.012$ & $0.500 \pm 0.497$
\\
& MPS             &     $0.211 \pm 0.019$ & $0.080 \pm 0.017$ & $0.100 \pm 0.046$
 \\
& GS                     & $0.212 \pm 0.026$ & $0.077 \pm 0.018$ & $0.096 \pm 0.035$

\\
\hline
\multirow{5}{*}{STL2 (Always Stable)} 
& -                     &$0.170 \pm 0.071$ & $0.113 \pm 0.011$ & $0.075 \pm 0.040$

 \\
& CMS            &    $0.153 \pm 0.021$ & $0.126 \pm 0.013$ & $0.074 \pm 0.030$

\\
& EDS                     & $0.114 \pm 0.063$ & $0.114 \pm 0.015$ &\textbf{ 0.056 ± 0.028}

\\
& MPS             &    $0.188 \pm 0.009$ & $0.118 \pm 0.015$ & $0.063 \pm 0.013$

 \\
& GS                     & $0.164 \pm 0.074$ & $0.111 \pm 0.013$ & $0.070 \pm 0.039$

\\
\hline
\multirow{5}{*}{STL3 (Locally Smooth) } 
& -                     &$0.184 \pm 0.021$ & $0.154 \pm 0.015$ & $0.099 \pm 0.031$

 \\
& CMS            &    $0.122 \pm 0.018$ & $0.081 \pm 0.037$ & $0.084 \pm 0.030$

\\
& EDS                     & $0.149 \pm 0.008$ & \textbf{0.076 ± 0.040} & $0.083 \pm 0.034$

\\
& MPS             &     $0.171 \pm 0.025$ & $0.151 \pm 0.030$ & $0.083 \pm 0.046$

 \\
& GS                     & $0.180 \pm 0.017$ & $0.118 \pm 0.006$ & $0.091 \pm 0.043$

\\

\hline
\end{tabular}
\caption{Expected Calibration Error (ECE) comparison across confidence sources and estimation strategies. STL-based methods, particularly STL1–STL3 combined with Uncertainty Reshaping (CMS, EDS), consistently yield better calibration than traditional techniques.}
\label{tab:ece_results}

\end{table*}


\noindent \textbf{Experimental Setup:} We conduct our experiments using the {Qwen-7B} language model,\footnote{\url{https://huggingface.co/Qwen/Qwen-7B}} a high-performing open-source LLM optimized for Chinese and mathematical reasoning tasks. Our evaluation is conducted on all multiple-choice questions from Chinese national college entrance exams ({Gaokao}) spanning 2010 to 2022, totaling 12 years of official high school mathematics problems. These questions are drawn from {GAOKAO-Bench}~\cite{zhang2023evaluating}, to assess LLMs' language understanding and symbolic reasoning capabilities using real-world exam data. Table~\ref{tab:math_examples} illustrates representative examples.

To prevent the model from relying on surface-level pattern matching or memorized templates, we augment the dataset following strategies inspired by GSM-Symbolic~\cite{mirzadeh2024gsm}, which shows that LLMs often fail when symbols, numbers, or phrasing are changed. We use a more advanced reasoning model -- {OpenAI's {o1} API} to perform all paraphrasing operations in a controlled and semantically faithful manner. Specifically, each original problem is rewritten to preserve logical structure and correct answer while varying lexical expressions (e.g., transforming \texttt{"find the intersection of sets A and B"} into \texttt{"determine the elements shared by both A and B"}). We further introduce linguistic variation through backtranslation, translating problems into a pivot language (such as French) and then back to English, thereby injecting natural noise without changing semantics. Additionally, symbolic formulations are diversified using template-based transformations—for example, the set expression ``\texttt{\(A \cap B\)}'' may be rephrased as ``\texttt{\(x \in {Z}, \sqrt{x} \leq 4\)}'' or ``\texttt{\(B = \{x \in {Z} \mid x^2 \leq 16\}\)}''. These augmentations collectively evaluate the model’s robustness to paraphrasing, symbol rewriting, and structural variation, ensuring assessment focuses on genuine reasoning rather than memorized syntax. The correct answer for each augmented sample is inherited from the corresponding original Gaokao-Bench problem, since paraphrasing and symbolic transformations preserve semantic and logical equivalence. In total, the original 432 questions are each rewritten twice, resulting in a final dataset of 1,296 problem instances for evaluation.

\noindent \textbf{Quantitative Results and Analysis:} To illustrate how different estimation methods affect calibration behavior, Figure~\ref{fig:results} shows an example confidence histogram under four representative strategies. While it only reflects a single problem instance and one STL constraint (STL1), the figure demonstrates how reshaping and temporal logic evaluation yield more aligned and interpretable confidence estimates.
We now turn to aggregate results across the full test set. All scores are averaged over three runs, and $\pm$ denotes standard deviation caused by the randomness introduced through temperature-controlled decoding, which affects the variability and creativity of LLM outputs. Table~\ref{tab:ece_results} presents the {Expected Calibration Error (ECE)} for various estimation strategies across three types of uncertainty sources: logits-based, self-evaluation-based, and internal-based. Table~\ref{tab:brier_results} complements this with {Brier Scores}, which jointly capture calibration and sharpness of probabilistic estimates.

From the ECE results, we observe that traditional post-hoc calibration methods such as Temperature Scaling~\cite{guo2017calibration} and Histogram Binning~\cite{zadrozny2001obtaining} reduce miscalibration compared to raw one-step uncertainty. For example, Histogram Binning achieves an ECE of $0.139$ on logits-based predictions, improving substantially over the one-step baseline ($0.324$). However, these methods operate globally and do not account for the multi-step nature of reasoning in LLMs.

In contrast, CoT-based methods yield stronger performance, especially when combined with our proposed Uncertainty Reshaping Strategies. Simply averaging confidence across CoT steps reduces ECE across all sources, and applying smoothing techniques such as Causal Minimum Smoothing (CMS) or Exponential Decay Smoothing (EDS) brings further gains. For instance, CMS reduces logits-based ECE to $0.107$, the lowest among all non-STL methods.

STL-based temporal evaluation further improves calibration. By enforcing high-level temporal constraints like \textit{Eventually Confident} (STL1), \textit{Always Stable} (STL2), and \textit{Locally Smooth} (STL3), the model's confidence trajectory becomes more interpretable and aligned with reasoning quality. STL1 combined with GS achieves an ECE of $0.077$ on self-evaluation-based confidence and $0.096$ on internal-based, outperforming all other approaches. Notably, STL2 with EDS reaches an ECE of $0.056$ on internal-based confidence—the best result across all settings.

The Brier Score analysis mirrors this trend. STL methods consistently produce lower scores than CoT average or standard post-hoc techniques. STL1 with CMS achieves the best self-evaluation-based Brier Score ($0.234$), while STL2 with EDS offers the best internal-based score ($0.056$). These results confirm that applying STL logic not only improves calibration error but also leads to sharper, more reliable probability estimates.

Overall, STL-based confidence estimation outperforms traditional calibration and CoT-only baselines, particularly when paired with CMS or EDS. These findings highlight the value of structured temporal logic as a calibration framework for LLM-based reasoning, offering both theoretical guarantees and empirical gains.

\begin{table*}[h]
\centering
\small
\begin{tabular}{|c|l|c|c|c|}
\hline
\textbf{Method} & \textbf{ Reshaping Strategy} & \textbf{Logits-based $\downarrow$} & \textbf{Self-evaluation-based  $\downarrow$} & \textbf{Internal-based $\downarrow$} \\
\hline
 1-step Uncertainty&- & $0.339 \pm 0.019$ & $0.677 \pm 0.032$ & $0.678 \pm 0.032$
\\
\hline

Temperature Scaling&- &  $0.263 \pm 0.040$ & $0.255 \pm 0.018$ & $0.276 \pm 0.018$

\\
\hline
Histogram Binning&- & $0.284 \pm 0.002$ & $0.263 \pm 0.020$ & $0.259 \pm 0.021$

\\
\hline
\multirow{5}{*}{CoT Average} &- & $0.225 \pm 0.033$ & $0.498 \pm 0.023$ & $0.564 \pm 0.028$
\\
&CMS&$0.218 \pm 0.032$ & $0.442 \pm 0.019$ & $0.537 \pm 0.029$

\\

&EDS&$0.219 \pm 0.032$ & $0.455 \pm 0.017$ & $0.530 \pm 0.027$

\\

&MPS& $0.219 \pm 0.033$ & $0.483 \pm 0.022$ & $0.563 \pm 0.028$

\\

&GS& $0.222 \pm 0.033$ & $0.493 \pm 0.022$ & $0.536 \pm 0.037$

\\
\hline
\multirow{5}{*}{STL1 (Eventually Confident)} 
& -             &      $0.223 \pm 0.003$ & $0.244 \pm 0.001$ & $0.246 \pm 0.001$
 \\
& CMS            &   \textbf{0.218 ± 0.005} & \textbf{0.234 ± 0.000} & $0.240 \pm 0.000$

\\
& EDS                     & $0.219 \pm 0.003$ & $0.238 \pm 0.001$ & $0.241 \pm 0.001$
\\
& MPS             &     $0.221 \pm 0.000$ & $0.243 \pm 0.002$ & $0.246 \pm 0.001$
 \\
& GS                     & $0.223 \pm 0.004$ & $0.243 \pm 0.002$ & $0.239 \pm 0.001$

\\
\hline
\multirow{5}{*}{STL2 (Always Stable)} 
& -                     &$0.238 \pm 0.008$ & $0.252 \pm 0.005$ & $0.254 \pm 0.002$

 \\
& CMS            &   $0.239 \pm 0.009$ & $0.246 \pm 0.002$ & $0.249 \pm 0.001$

\\
& EDS                     & $0.243 \pm 0.004$ & $0.253 \pm 0.003$ & $0.253 \pm 0.001$

\\
& MPS             &    $0.239 \pm 0.006$ & $0.252 \pm 0.004$ & $0.254 \pm 0.002$

 \\
& GS                     & $0.238 \pm 0.008$ & $0.252 \pm 0.004$ & $0.257 \pm 0.005$

\\
\hline
\multirow{5}{*}{STL3 (Locally Smooth) } 
& -                     &$0.232 \pm 0.011$ & $0.236 \pm 0.002$ & $0.240 \pm 0.001$

 \\
& CMS            &    $0.237 \pm 0.008$ & $0.241 \pm 0.002$ & $0.242 \pm 0.001$

\\
& EDS                     &$0.239 \pm 0.006$ & $0.242 \pm 0.001$ & $0.244 \pm 0.000$

\\
& MPS             &    $0.236 \pm 0.010$ & $0.236 \pm 0.003$ & \textbf{0.238 ± 0.002}

 \\
& GS                     & $0.233 \pm 0.011$ & $0.235 \pm 0.001$ & $0.249 \pm 0.002$

\\

\hline
\end{tabular}
\caption{Brier Score comparison across methods. Lower scores indicate better-calibrated and sharper probabilistic predictions. STL-based temporal constraints further improve performance beyond CoT averaging and post-hoc calibration.}
\label{tab:brier_results}

\end{table*}

\section{Conclusion}

This paper presents a structured approach to confidence estimation for LLM-based mathematical reasoning. By modeling stepwise confidence as a temporal signal and evaluating its quality using STL, our method addresses limitations in traditional calibration and step-level aggregation techniques. We introduce a suite of uncertainty reshaping strategies and STL-based robustness constraints that enforce desirable properties such as eventual certainty, monotonic progression, and local smoothness.

Experimental results on GAOKAO-Bench demonstrate that our STL-based evaluation, particularly when paired with smoothing strategies like CMS and EDS, consistently achieves lower ECE and Brier Scores compared to standard baselines. Beyond quantitative improvements, the framework provides a principled, interpretable method for diagnosing and enhancing reasoning quality in educational LLM applications.

\section{Limitations}

While our method improves calibration and interpretability, it is currently limited to high school-level multiple-choice math problems. Extending the framework to open-ended questions, formal proofs, or multi-modal reasoning is a promising direction.
Since all experiments are conducted on Qwen-7B, generalization to other models remains uncertain. Testing on models like Gemma 3, Llama 3.2, or DeepSeek would help assess robustness across architectures.

Both the reshaping strategies and STL specifications are manually defined. Future work could explore learning them dynamically via reinforcement learning or differentiable logic, enabling more adaptive and data-driven calibration beyond manual tuning.

This paper focuses on linear chain-of-thought reasoning. More complex prompting paradigms, such as tree-of-thought~\cite{yao2023tree}, introduce branching and cyclic structures that pose new challenges for temporal modeling. Extending our evaluation to computation tree logic (CTL) to accommodate such structures would broaden its applicability to richer and more realistic cognitive processes.

Finally, our method is post-hoc and does not influence model inference. Integrating uncertainty feedback into real-time tutoring systems could enable dynamic intervention, hinting, and early error detection.

\bibliography{anthology,custom}

\begin{thebibliography}{30}
\expandafter\ifx\csname natexlab\endcsname\relax\def\natexlab#1{#1}\fi

\bibitem[{Ben{\'\i}tez et~al.(2024)Ben{\'\i}tez, Xu, Boudreau, Kow, Bello, Van~Phuoc, Wang, Sun, Leung, Lan et~al.}]{benitez2024harnessing}
Trista~M Ben{\'\i}tez, Yueyuan Xu, J~Donald Boudreau, Alfred Wei~Chieh Kow, Fernando Bello, Le~Van~Phuoc, Xiaofei Wang, Xiaodong Sun, Gilberto Ka-Kit Leung, Yanyan Lan, et~al. 2024.
\newblock Harnessing the potential of large language models in medical education: promise and pitfalls.
\newblock \emph{Journal of the American Medical Informatics Association}, 31(3):776--783.

\bibitem[{Desai and Durrett(2020)}]{desai2020calibration}
Shrey Desai and Greg Durrett. 2020.
\newblock Calibration of pre-trained transformers.
\newblock \emph{arXiv preprint arXiv:2003.07892}.

\bibitem[{Diao et~al.(2023)Diao, Wang, Lin, Pan, Liu, and Zhang}]{diao2023active}
Shizhe Diao, Pengcheng Wang, Yong Lin, Rui Pan, Xiang Liu, and Tong Zhang. 2023.
\newblock Active prompting with chain-of-thought for large language models.
\newblock \emph{arXiv preprint arXiv:2302.12246}.

\bibitem[{Donz{\'e} and Maler(2010)}]{donze2010robust}
Alexandre Donz{\'e} and Oded Maler. 2010.
\newblock Robust satisfaction of temporal logic over real-valued signals.
\newblock In \emph{International Conference on Formal Modeling and Analysis of Timed Systems}, pages 92--106. Springer.

\bibitem[{Fainekos and Pappas(2006)}]{fainekos2006robustness}
Georgios~E Fainekos and George~J Pappas. 2006.
\newblock Robustness of temporal logic specifications.
\newblock In \emph{International Workshop on Formal Approaches to Software Testing}, pages 178--192. Springer.

\bibitem[{Gan et~al.(2023)Gan, Qi, Wu, and Lin}]{gan2023large}
Wensheng Gan, Zhenlian Qi, Jiayang Wu, and Jerry Chun-Wei Lin. 2023.
\newblock Large language models in education: Vision and opportunities.
\newblock In \emph{2023 IEEE international conference on big data (BigData)}, pages 4776--4785. IEEE.

\bibitem[{Guo et~al.(2017)Guo, Pleiss, Sun, and Weinberger}]{guo2017calibration}
Chuan Guo, Geoff Pleiss, Yu~Sun, and Kilian~Q Weinberger. 2017.
\newblock On calibration of modern neural networks.
\newblock In \emph{International conference on machine learning}, pages 1321--1330. PMLR.

\bibitem[{Idris et~al.(2024)Idris, Feng, and Dyo}]{idris2024revolutionising}
Mohamed~Diab Idris, Xiaohua Feng, and Vladimir Dyo. 2024.
\newblock Revolutionising higher education: Unleashing the potential of large language models for strategic transformation.
\newblock \emph{IEEE Access}.

\bibitem[{Jiang et~al.(2021)Jiang, Araki, Ding, and Neubig}]{jiang2021can}
Zhengbao Jiang, Jun Araki, Haibo Ding, and Graham Neubig. 2021.
\newblock How can we know when language models know? on the calibration of language models for question answering.
\newblock \emph{Transactions of the Association for Computational Linguistics}, 9:962--977.

\bibitem[{Kadavath et~al.(2022)Kadavath, Conerly, Askell, Henighan, Drain, Perez, Schiefer, Hatfield-Dodds, DasSarma, Tran-Johnson et~al.}]{kadavath2022language}
Saurav Kadavath, Tom Conerly, Amanda Askell, Tom Henighan, Dawn Drain, Ethan Perez, Nicholas Schiefer, Zac Hatfield-Dodds, Nova DasSarma, Eli Tran-Johnson, et~al. 2022.
\newblock Language models (mostly) know what they know.
\newblock \emph{arXiv preprint arXiv:2207.05221}.

\bibitem[{Kasneci et~al.(2023)Kasneci, Se{\ss}ler, K{\"u}chemann, Bannert, Dementieva, Fischer, Gasser, Groh, G{\"u}nnemann, H{\"u}llermeier et~al.}]{kasneci2023chatgpt}
Enkelejda Kasneci, Kathrin Se{\ss}ler, Stefan K{\"u}chemann, Maria Bannert, Daryna Dementieva, Frank Fischer, Urs Gasser, Georg Groh, Stephan G{\"u}nnemann, Eyke H{\"u}llermeier, et~al. 2023.
\newblock Chatgpt for good? on opportunities and challenges of large language models for education.
\newblock \emph{Learning and individual differences}, 103:102274.

\bibitem[{Li et~al.(2017)Li, Vasile, and Belta}]{li2017reinforcement}
Xiao Li, Cristian-Ioan Vasile, and Calin Belta. 2017.
\newblock Reinforcement learning with temporal logic rewards.
\newblock In \emph{2017 IEEE/RSJ International Conference on Intelligent Robots and Systems (IROS)}, pages 3834--3839. IEEE.

\bibitem[{Lyu et~al.(2025)Lyu, Shridhar, Malaviya, Zhang, Elazar, Tandon, Apidianaki, Sachan, and Callison-Burch}]{lyu2025calibrating}
Qing Lyu, Kumar Shridhar, Chaitanya Malaviya, Li~Zhang, Yanai Elazar, Niket Tandon, Marianna Apidianaki, Mrinmaya Sachan, and Chris Callison-Burch. 2025.
\newblock Calibrating large language models with sample consistency.
\newblock In \emph{Proceedings of the AAAI Conference on Artificial Intelligence}, volume~39, pages 19260--19268.

\bibitem[{Mirzadeh et~al.(2024)Mirzadeh, Alizadeh, Shahrokhi, Tuzel, Bengio, and Farajtabar}]{mirzadeh2024gsm}
Iman Mirzadeh, Keivan Alizadeh, Hooman Shahrokhi, Oncel Tuzel, Samy Bengio, and Mehrdad Farajtabar. 2024.
\newblock Gsm-symbolic: Understanding the limitations of mathematical reasoning in large language models.
\newblock \emph{arXiv preprint arXiv:2410.05229}.

\bibitem[{Naeini et~al.(2015)Naeini, Cooper, and Hauskrecht}]{naeini2015obtaining}
Mahdi~Pakdaman Naeini, Gregory Cooper, and Milos Hauskrecht. 2015.
\newblock Obtaining well calibrated probabilities using bayesian binning.
\newblock In \emph{Proceedings of the AAAI conference on artificial intelligence}, volume~29.

\bibitem[{Polyxeni Paulina~Kastania(2024)}]{polyxeni2024building}
Nikoleta Polyxeni Paulina~Kastania. 2024.
\newblock Building trust in ai education: Addressing transparency and ensuring.
\newblock \emph{Trust and Inclusion in AI-mediated Education: Where Human Learning Meets Learning Machines}, page~73.

\bibitem[{Rescher and Urquhart(2012)}]{rescher2012temporal}
Nicholas Rescher and Alasdair Urquhart. 2012.
\newblock \emph{Temporal logic}, volume~3.
\newblock Springer Science \& Business Media.

\bibitem[{Shah et~al.(2024)Shah, Yu, Lyu, Park, Yu, He, Ke, Mozer, Bengio, Arora et~al.}]{shah2024ai}
Vedant Shah, Dingli Yu, Kaifeng Lyu, Simon Park, Jiatong Yu, Yinghui He, Nan~Rosemary Ke, Michael Mozer, Yoshua Bengio, Sanjeev Arora, et~al. 2024.
\newblock Ai-assisted generation of difficult math questions.
\newblock \emph{arXiv preprint arXiv:2407.21009}.

\bibitem[{Tanneru et~al.(2024)Tanneru, Agarwal, and Lakkaraju}]{tanneru2024quantifying}
Sree~Harsha Tanneru, Chirag Agarwal, and Himabindu Lakkaraju. 2024.
\newblock Quantifying uncertainty in natural language explanations of large language models.
\newblock In \emph{International Conference on Artificial Intelligence and Statistics}, pages 1072--1080. PMLR.

\bibitem[{van Huijgevoort et~al.(2024)van Huijgevoort, Wang, Soudjani, and Haesaert}]{van2024specification}
Birgit~C van Huijgevoort, Ruohan Wang, Sadegh Soudjani, and Sofie Haesaert. 2024.
\newblock Specification-guided temporal logic control for stochastic systems: a multi-layered approach.
\newblock \emph{arXiv preprint arXiv:2407.03896}.

\bibitem[{Wang et~al.(2024)Wang, Xu, Li, Zhang, Liang, Tang, Yu, and Wen}]{wang2024large}
Shen Wang, Tianlong Xu, Hang Li, Chaoli Zhang, Joleen Liang, Jiliang Tang, Philip~S Yu, and Qingsong Wen. 2024.
\newblock Large language models for education: A survey and outlook.
\newblock \emph{arXiv preprint arXiv:2403.18105}.

\bibitem[{Wang et~al.(2022)Wang, Wei, Schuurmans, Le, Chi, Narang, Chowdhery, and Zhou}]{wang2022self}
Xuezhi Wang, Jason Wei, Dale Schuurmans, Quoc Le, Ed~Chi, Sharan Narang, Aakanksha Chowdhery, and Denny Zhou. 2022.
\newblock Self-consistency improves chain of thought reasoning in language models.
\newblock \emph{arXiv preprint arXiv:2203.11171}.

\bibitem[{Wei et~al.(2022)Wei, Wang, Schuurmans, Bosma, Xia, Chi, Le, Zhou et~al.}]{wei2022chain}
Jason Wei, Xuezhi Wang, Dale Schuurmans, Maarten Bosma, Fei Xia, Ed~Chi, Quoc~V Le, Denny Zhou, et~al. 2022.
\newblock Chain-of-thought prompting elicits reasoning in large language models.
\newblock \emph{Advances in neural information processing systems}, 35:24824--24837.

\bibitem[{Yao et~al.(2023)Yao, Yu, Zhao, Shafran, Griffiths, Cao, and Narasimhan}]{yao2023tree}
Shunyu Yao, Dian Yu, Jeffrey Zhao, Izhak Shafran, Tom Griffiths, Yuan Cao, and Karthik Narasimhan. 2023.
\newblock Tree of thoughts: Deliberate problem solving with large language models.
\newblock \emph{Advances in neural information processing systems}, 36:11809--11822.

\bibitem[{Zadrozny and Elkan(2001)}]{zadrozny2001obtaining}
Bianca Zadrozny and Charles Elkan. 2001.
\newblock Obtaining calibrated probability estimates from decision trees and naive bayesian classifiers.
\newblock In \emph{Icml}, volume~1.

\bibitem[{Zhang et~al.(2023)Zhang, Li, Zong, Ying, He, and Qiu}]{zhang2023evaluating}
Xiaotian Zhang, Chunyang Li, Yi~Zong, Zhengyu Ying, Liang He, and Xipeng Qiu. 2023.
\newblock Evaluating the performance of large language models on gaokao benchmark.
\newblock \emph{arXiv preprint arXiv:2305.12474}.

\bibitem[{Zhang et~al.(2022)Zhang, Zhang, Li, and Smola}]{zhang2022automatic}
Zhuosheng Zhang, Aston Zhang, Mu~Li, and Alex Smola. 2022.
\newblock Automatic chain of thought prompting in large language models.
\newblock \emph{arXiv preprint arXiv:2210.03493}.

\bibitem[{Zhao et~al.(2021)Zhao, Wallace, Feng, Klein, and Singh}]{zhao2021calibrate}
Zihao Zhao, Eric Wallace, Shi Feng, Dan Klein, and Sameer Singh. 2021.
\newblock Calibrate before use: Improving few-shot performance of language models.
\newblock In \emph{International conference on machine learning}, pages 12697--12706. PMLR.

\bibitem[{Zhou et~al.(2025)Zhou, Zhang, Jiang, Gao, Liu, and Jiang}]{zhou2025study}
Yizhou Zhou, Mengqiao Zhang, Yuan-Hao Jiang, Xinyu Gao, Naijie Liu, and Bo~Jiang. 2025.
\newblock A study on educational data analysis and personalized feedback report generation based on tags and chatgpt.
\newblock \emph{arXiv preprint arXiv:2501.06819}.

\bibitem[{Zhu et~al.(2025)Zhu, Li, Jiang, Li, Mei, Jin, and Dong}]{zhu2025uncertainty}
Yuqi Zhu, Ge~Li, Xue Jiang, Jia Li, Hong Mei, Zhi Jin, and Yihong Dong. 2025.
\newblock Uncertainty-guided chain-of-thought for code generation with llms.
\newblock \emph{arXiv preprint arXiv:2503.15341}.

\end{thebibliography}
\bibliographystyle{acl_natbib}




\end{document}